\documentclass{article}
\usepackage{spconf,amsmath,epsfig}
\usepackage{multirow}
\usepackage{subfigure}
\usepackage{amssymb}

\title{Self-supervised Spectral Matching Network for Hyperspectral Target Detection}
\name{Can Yao, Yuan Yuan, Zhiyu Jiang\sthanks{2021 IEEE. Personal use of this material is permitted. Permission from IEEE must be obtained for all other uses, in any current or future media, including reprinting/republishing this material for advertising or promotional purposes, creating new collective works, for resale or redistribution to servers or lists, or reuse of any copyrighted component of this work in other works. Corresponding author: Zhiyu Jiang (jiangzhiyu@nwpu.edu.cn).}}
\address{School of Computer Science and School of Artificial Intelligence, Optics and Electronics (iOPEN), \\
	Northwestern Polytechnical University, Xi'an 710072, P.R. China}
\begin{document}
%
\maketitle
\begin{abstract}
Hyperspectral target detection is a pixel-level recognition problem. Given a few target samples, it aims to identify the specific target pixels such as airplane, vehicle, ship, from the entire hyperspectral image.
In general, the background pixels take the majority of the image and complexly distributed.
As a result, the datasets are weak annotated and extremely imbalanced.
To address these problems, a spectral mixing based self-supervised paradigm is designed for hyperspectral data to obtain an effective feature representation.
The model adopts a spectral similarity based matching network framework. 
In order to learn more discriminative features, a pair-based loss is adopted to minimize the distance between target pixels while maximizing the distances between target and background.
Furthermore, through a background separated step, the complex unlabeled spectra are downsampled into different sub-categories. 
The experimental results on three real hyperspectral datasets demonstrate that the proposed framework achieves better results compared with the existing detectors.

\end{abstract}
\vspace{-0.3em}
\section{Introduction}
\label{sec:intro}
\vspace{-0.5em}

Hyperspectral technology plays an important role in remote sensing\cite{WangCGCCTJ19,bioucas2013hyperspectral}.
Hyperspectral imagery (HSI) includes both spatial information and spectral information.
Due to the abundant information contained in the spectrum, hyperspectral imagery has great advantages in detecting small targets with fewer pixels.
Target detection in hyperspectral images is a very practical and significant research field \cite{Chang2005Orthogonal,WangBGH19}.
Different from object detection, hyperspectral target detection is a pixel-level one-class classification problem.
Given a few target pixels, it aims to identify all target pixels from the entire image \cite{lu2017hybrid}. 
Although HSI has abundant spectral information, it is quite a challenging task since hundreds of spectrum dimensionalities tend to generate data redundancy and increase calculation consumption dramatically.
The targets are usually very small and surrounded by large and complex background, let alone some background pixels are very similar to the target. 
Worse still, the images are weakly annotated because only a small number of target pixels are given. Considering that the background takes the majority part of the image, the target and background pixels are extremely imbalanced. 

There is a stable development of hyperspectral target detection in the literature \cite{manolakis2003hyperspectral,Chang2000Generalized,zhu2019target}. 
Earlier works adopt the spectrum as the representation of pixels directly and detect the target pixels by simple signal processing or sparse coding algorithms \cite{Chen2011Sparse,li2015combined}. Recently, inspired by the great success of deep learning in computer vision tasks, deep neural networks have been employed to learn more discriminative representations\cite{shi2019discriminative}.
Although better performance is achieved, these approaches also suffer from the problems of lacking annotated data as well as extremely sample imbalance. 
Some of them adopt a concentric windows method to combine the local background pixels and given target pixels.
The center pixels which have a larger difference with surroundings and similar to given targets would be considered as a target. 
However, this method is based on the hypothesis that all targets are very small, most of the target areas are included in the inner window, while most of the pixels contained in the outer window are background.
If the targets are complex, or there are a large number of interference samples which similar to the targets, this hypothesis will be affected. 
There are also some methods to manually label a variety of background samples.
Hyperspectral images are difficult to label the pixels’ category by naked eyes, which is time-consuming and hard to be widely used.

\begin{figure*}[htbp]

	\begin{minipage}[b]{1.0\linewidth}
		\includegraphics[width=17.7cm]{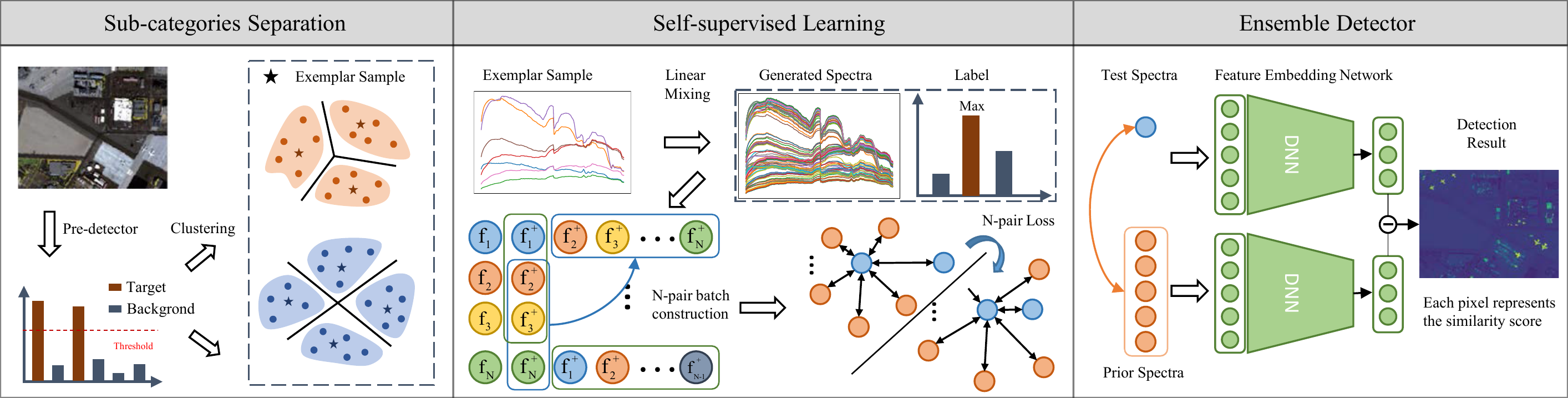}
	\end{minipage}	
	\vspace{-2em}
	\caption{Flowchart of the proposed method. The pre-detector roughly separates the spectra into target and background categories. Then the separated results are clustering into different sub-categories. 
	Mixing exemplars of each cluster by linear combination, the new spectra can be generated.
	Using the generated data, the encoder network can be trained through the pretext task.
	And then, the encoder network is trained by the $N$-pair loss to learn more discriminative features. Finally, the detection score is calculated by the ensemble similarity with each prior target.}
	\label{fig:flow}
	\vspace{-1.2em}
\end{figure*}
In this paper, to better discriminate target and background, we adopt a coarse-grained classification step to separate the unlabeled data into different sub-categories. 
First, we roughly select a target and background set by a pre-detector. 
According to the given target, pre-detector can suppress the background spectra, and target samples would obtain a higher score. 
With the result score of pre-detector, a clustering method is adopted to the target and background data. The whole data are divided into different sub-categories. 
Therefore, the pseudo labels can be obtained for the unlabeled data to model the complex background samples. 
Next, the centroids of pseudo-labeled data are selected to generate large amounts of training data that can be trained in a 
supervised manner. 
Through such a pretext task, we can learn a more discriminative feature representation through a self-supervised paradigm. 
Finally, we train the feature embedding network with a pair-based loss which can minimize the distance between target pixels and maximize the distances between target and background. 
With a hard negative sample mining strategy, a model with strong discrimination ability will be obtained. By calculating the similarity with the given target, we can obtain the specific target location's in the hyperspectral image.


The contributions of this work can be summarized as follows:

\begin{itemize}	
\setlength{\itemsep}{-0.1em}
\item A coarse-grained classification strategy is adopted to classify the unlabeled samples into different sub-categories. 
This strategy can alleviate the problem of sample imbalance and the variability of the background spectra. 
\item 
A self-supervised learning method based on spectral mixing characteristics is proposed.
In the case of limited annotation data, the feature extractor network can learn a more discriminative feature representation in a self-supervised fashion. 
\item Our work presents a novel spectral similarity based matching framework for hyperspectral target detection.
The experimental results on three real hyperspectral datasets verify the effectiveness of our method.
\end{itemize}

%

\section{PROPOSED DETECTION FRAMEWORK}
The proposed target detection method mainly includes the following steps: the coarse-grained classification, self-supervised feature representation learning, and the final detector.
The whole process can be seen from Fig. \ref{fig:flow}.


\vspace{-1em}
\subsection{Coarse-grained Classification with Sub-categories Separation}
\vspace{-0.3em}
The spectra can be roughly classified into target and background category through a simple pre-detector. 
The constrained energy minimization (CEM) \cite{Chang2000Generalized} detector is adopted as a coarse-grained detector.

Consider all pixels of hyperspectral dataset as  
$\mathbf{H}= \left \{ \mathbf{h}_{i} \right \}_{i=1}^{N} \in \mathbb{R}^{N \times B}$, where $N$ is the number of pixels, and $B$ is the number of spectral bands. $\mathbf{h}_{i} \in \mathbb{R}^{B \times 1}$ represent the pixel in HSI.
For a linear filter  $\mathbf{w} = \left [ w _{1},w _{2},\cdots ,w _{B} \right ]^\mathrm{T}$ and a given target pixels $\mathbf{d}$, the CEM detector setting $\mathbf{w} ^\mathrm{T} \mathbf{d}= 1$ while suppressing the background pixel output.
The CEM detector is formulated as follows,
\begin{equation}
\setlength{\abovedisplayskip}{3pt} 
\setlength{\belowdisplayskip}{3pt}
D_{\mathrm{CEM}}(\mathbf{h})= \mathbf{w}^\mathrm{T} \mathbf{h}= \frac{\mathbf{h}^\mathrm{T} \mathbf{R}^{-1} \mathbf{d}}{\mathbf{d}^\mathrm{T} \mathbf{R}^{-1} \mathbf{d}},
\end{equation}
where $\mathbf{R}=(1 / N) \sum\limits_{i=1}^{N} \mathbf{h}_{i} \mathbf{h}_{i}^\mathrm{T}$ is a autocorrelation matrix.

By ranking the results of CEM detector, 1\% of the samples are excluded as target set $\mathbf{H}_{t}$ and the rests are purified background set $\mathbf{H}_{b}$. 
Practically, the CEM detector is influenced by the variance of the given pixel. To enhance the robustness, we take the average of $D_{\mathrm{CEM}}(\mathbf{h})$ of the multiple targets pixels as the final detection results.

As the excessive number of background samples, this leads to the extreme imbalance of training data. 
In addition, due to the varity and complexity of spectra, there are some differences within background or target samples. 
In order to resample the data, we apply a k-means clustering method to the background set $\mathbf{H}_{b}$  and target set $\mathbf{H}_{t}$ respectively based on the result of CEM pre-detector.
This method aims to minimize the following objective function:
\begin{equation}
\setlength{\abovedisplayskip}{3pt} 
\setlength{\belowdisplayskip}{-3pt}
J=\sum_{j=1}^{K} \sum_{i=1}^{N}\left\|\mathbf{x}_{i}^{j}-\mathbf{c}_{j}\right\|^{2},
\vspace{-1pt}
\end{equation}
where $\left\|\mathbf{x}_{i}^{j}-\mathbf{c}_{j}\right\|^{2}$ indicates the Euclidean distance between the data point $\mathbf{x}_{i}^{j}$ and the cluster center $\mathbf{c}_{j}$. $K$ is the number of cluster.

\vspace{-0.3em}
\subsection{Self-supervised Representation Learning with Pair-based Loss}
\vspace{-0.3em}
To make full use of the unlabeled data, we train the model in a self-supervised manner.
According to spectral mixing characteristics, we design a pretext task for spectral data. For cluster centers $\mathbf{C} = \left \{ \mathbf{c}_{1}, \mathbf{c}_{2}, \mathbf{c}_{3},... ,\mathbf{c}_{K}\right \}$ where $K$ is the number of cluster centers, a large number of samples can be generated by linear combination of cluster centers. The weight of the linear combination can be generated by the following formula:
\begin{equation}
\setlength{\abovedisplayskip}{3pt} 
\setlength{\belowdisplayskip}{3pt}
\alpha_{i}=\frac{\exp \left(z_{i} / T\right)}{\sum_{j} \exp \left(z_{j} / T\right)},
\end{equation}
where $T$ is the scaling factor, $z_{i}$ is sampled from a uniform distribution between 0 and 1.
Therefore, the generated data set can be expressed as $\mathbf{H^{'}} = \boldsymbol{\alpha} \cdot \mathbf{C}$. 
For each $\mathbf{h}_{i}$, the label can be obtain by $ y_{i} = \mathop{\max}\left ( \boldsymbol{\alpha} \right )$.
Then the feature extractor can be trained with the generated data by a supervised method.
With the training process of pretext task, the embedded network can extract the features which contain spectral semantic information.

In order to separate targets and backgrounds, the model should be minimizing the distance between target pixels while maximizing the distances between target and background. Therefore, we adopt $N$-pair loss\cite{sohn2016improved} to constraint multiple subcategories simultaneously.
The training data sample from each class, composed of the tuplet data $\left\{\mathbf{x}, \mathbf{x}^{+}, \mathbf{x}_{1}, \cdots \mathbf{x}_{N-1}\right\}$, $\mathbf{x}^{+}$ is a positive sample of $\mathbf{x}$ and $\left\{\mathbf{x}_{i}\right\}_{i=1}^{N-1}$ is the negative samples.
The loss function is formulated as follows,

\begin{equation}
\setlength{\abovedisplayskip}{-4pt} 
\setlength{\belowdisplayskip}{3pt}
\mathcal{L}\left(\left\{\mathbf{x}, \mathbf{x}^{+},\mathbf{x}_{i}\right\}; f\right)=\log \left(1+\sum_{i=1}^{N-1} \exp \left(f^\mathrm{T} f_{i}-f^\mathrm{T} f^{+}\right)\right),
\end{equation}
where $f(\cdot ; \theta)$ is the embedding network.


\vspace{-0.3em}
\subsection{Ensemble Target Detector}
\vspace{-0.3em}
After the completion of the training process, the spectral matching network can measure the similarity from hyperspectral image pixels directly. Let $f(x)$ denotes the feature extraction network, 
the distance between the prior target pixel and each test sample can be calculated in the feature space.

For eliminating the influence by the variance of different given pixels, we adopt an ensemble method which considers the distance between multiple targets and sample to be tested.
For the given target pixel $\mathbf{H}_{t}=\left \{ \mathbf{h}_{ti} \right \}_{i=1}^{N_{t}}$, $N_{t}$ is the total number of given target, feeding each pixel to sub-network $f\left ( \mathbf{h}_{ti} \right )$, the original pixels are mapped to the feature space.
Then the similarity is calculated by euclidean distance, $ D_{i}= G\left ( f\left ( \mathbf{h} \right ),f\left ( \mathbf{h}_{ti} \right ) \right )$, and the final similarity score is computed as  
$S=\frac{1}{N_{t}}\left ( \sum_{i=1}^{N_{t}}D_{i}\right )$.

\begin{figure}[htbp]		
	\centering
	\includegraphics[width=8.5cm]{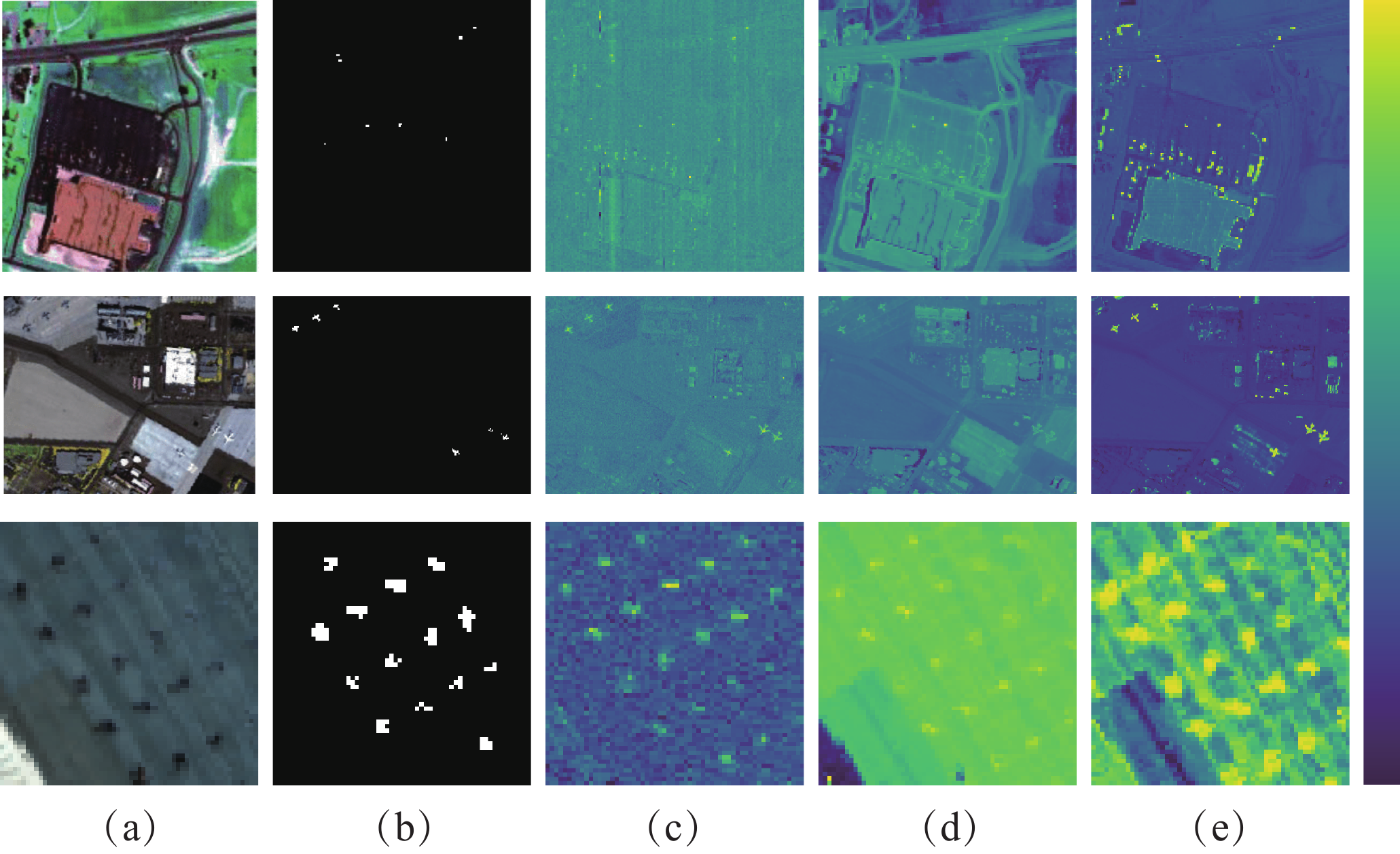}
	
	\caption{Results on three datasets. (a) False Color Image. (b) Ground Truth. (c) CEM. (d) CSCR. (e) Ours.}
	\vspace{-0em}
	\label{fig:vis}
\end{figure}

\section{Experimental Results and Analysis}
\vspace{-0.3em}
To verify the performance of our method, the proposed model was compared to the Adaptive Coherence Estimator (ACE)\cite{manolakis2003hyperspectral}, 
Constraint Energy Minimize (CEM)\cite{Chang2000Generalized}, 
Ensemble Constraint Energy Minimize (E-CEM)\cite{zhao2019ensemble}, 
Combined Sparse and Collaborative Representation (CSCR)\cite{li2015combined}, 
and DCSSAED\cite{shi2019discriminative}.
The three real hyperspectral datasets adopted are HYDICE, AVIRIS and AVIRIS2\cite{lu2017hybrid,zhu2019target}.

\vspace{-0.3em}
\subsection{Experimental Analysis}
\vspace{-0.3em}
In the experiment, 10 target spectra are randomly selected from the whole target spectra as the prior samples.
All other spectra are considered unlabeled.
With these prior spectra, we obtain possible background spectra from the entire image and construct pair data for supervised feature learning.
The feature embedding network contains one 1D convolution layer and two linear layers.
The input layer size of encoder network is set according to the spectral size of the dataset. The hidden layers are set as 128 and the feature embedding layer is set as 64. 
The training procedures are run with the SGD optimization algorithm with the initial learning rate of 0.0001 and batch size 128. 

\begin{figure}[ht]

	\centering
	\includegraphics[width=8.6cm]{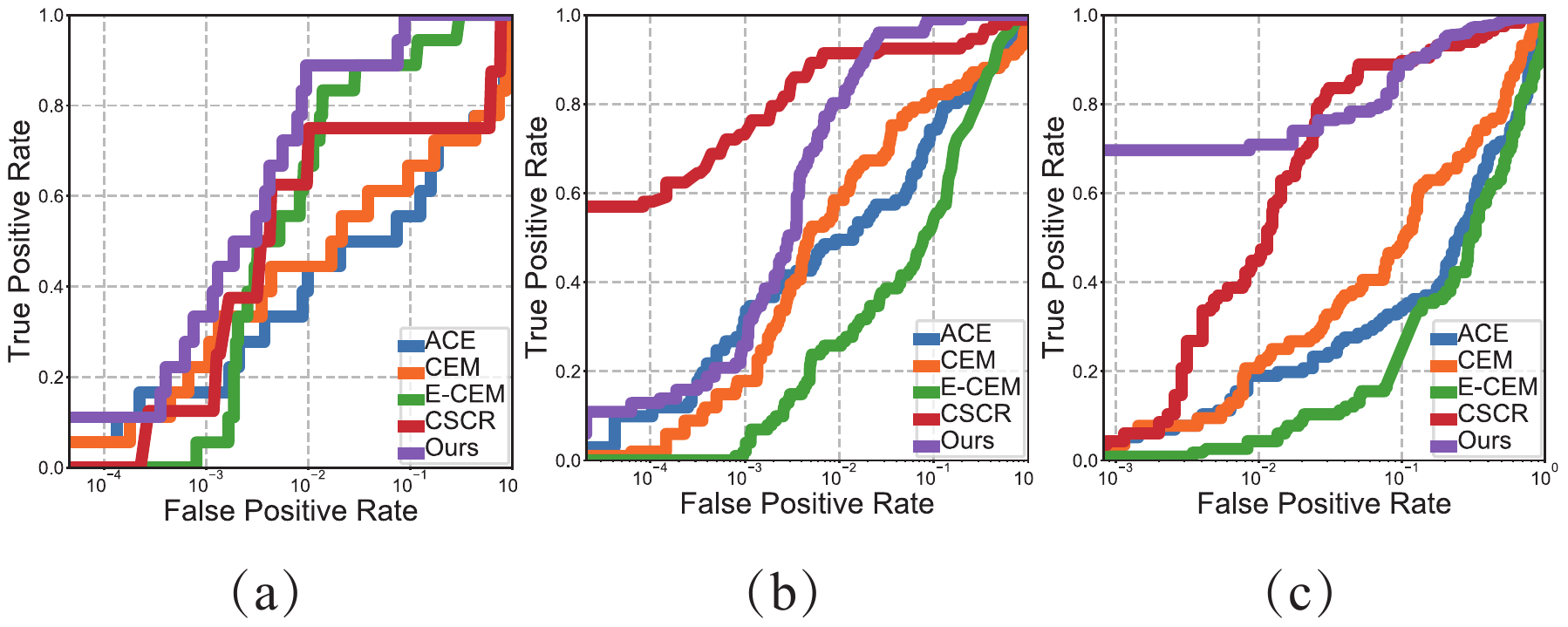}
	\vspace{-2.2em}
	\caption{ROC curves of different detectors on three datasets. (a) HYDICE. (b) AVIRIS. (c) AVIRIS2.}
	
	\label{fig:res}
\vspace{-2.2em}
\end{figure}

For HYDICE dataset,  it can be seen from Fig. \ref{fig:res}(a) the proposed  approach covers all other curves,  which indicates our method working better than all other methods.
From Table \ref{tab:cap} that the AUC value of our method are superior to other methods, which indicates that our method outperforms all the other detectors.
Furthermore, it can be seen from Fig. \ref{fig:vis}(e) that we can detect all the targets.
Moreover, while the target can be detected, the background noise still exists, but it is small relative to the whole sample size. It proves that the problem of imbalanced sample data is alleviated.

For AVIRIS and AVIRIS2 dataset, from Fig. \ref{fig:res}, our ROC curves intersect other curves. The perfomance is difficult to judge by the ROC curves.
However, it can be clearly seen from Table \ref{tab:cap} that we get a better AUC value, the performance of our method is superior to all the other detectors. 
In addition, it can be seen from Fig. \ref{fig:vis} that we can detect all the targets, with only a small number of false positives. 

Overall, from comparisons on the two datasets, we can simply find that: 1) By adopting the self-supervised learning, the more discriminative representations are learned. 2) Benefit from the $N$-pair loss, the sample imbalance problem is alleviated, so that the proposed approach can outperform other approaches. 
\vspace{-0.5em}

%
%
%
%
%
%

\begin{table}[tb]

	\begin{center}
		
		\caption{AUC results of different methods on three datasets.} \label{tab:cap}
		\renewcommand{\arraystretch}{1.2}
		\begin{tabular}{ccccc}
			\hline
			\multirow{2}{*}{Methods} & \multicolumn{2}{c}{Datasets}\\
			\cline{2-4}  
			
			&  HYDICE    &   AVIRIS  &   AVIRIS2 \\
			
			\hline
			
			ACE\cite{manolakis2003hyperspectral} & 0.7531 & 0.8121 & 0.6703 \\
			CEM\cite{Chang2000Generalized} & 0.9437 & 0.8730 & 0.7756 \\ 
			
			E-CEM\cite{Chang2005Orthogonal} & 0.7582 & 0.7934 & 0.6162  \\
			CSCR\cite{li2015combined} & 0.8249 & 0.9634 & 0.9513 \\
			DCSSAED\cite{shi2019discriminative} & 0.9649 & 0.9639  & 0.9549 \\
			
			Ours & \textbf{0.9882} & \textbf{0.9655} & \textbf{0.9573}\\
			\hline
		\end{tabular}
		
	\end{center}
	
\vspace{-2.2em}
\end{table}

\section{Conclusion}
\vspace{-0.3em}
In this work, a spectral similarity based matching network framework for hyperspectral target detection is proposed.
With the proposed structure, the effective and less redundant feature representations can be learned in an self-supervised manner.
In the embedded space, the distance of features between target pixels are minimized, while the distance of features between target and background are maximized by training with $N$-pair loss.
Experimental results on three real hyperspectral datasets demonstrate that the proposed framework achieves better results compared with the existing detectors. 
Due to the inner class difference between the spectra, the given prior spectra are different, the model results will be very different. 
This work adopts a simple ensemble method to alleviate this problem.
In further research, we will focus on the improvement of model robustness for hyperspectral target analysis.

\vspace{-0em}

%


\small 
\bibliographystyle{IEEEbib}
\bibliography{Template}
\end{document}